\documentclass[format=acmsmall, natbib=false, screen=true, nonacm=true]{acmart}

\RequirePackage[abbreviate=true, dateabbrev=true, isbn=true, doi=true, urldate=comp, url=true, maxbibnames=9, backref=false, backend=biber, style=ACM-Reference-Format, language=american]{biblatex}

\usepackage[utf8]{inputenc}
\usepackage[ruled]{algorithm2e} 
\usepackage{subfig}

\SetAlFnt{\small}
\SetAlCapFnt{\small}
\SetAlCapNameFnt{\small}
\SetAlCapHSkip{0pt}
\IncMargin{-\parindent}

\acmJournal{THRI}
\acmVolume{X}
\acmNumber{Y}
\acmYear{2019}
\acmMonth{2}
\copyrightyear{2019}

\setcopyright{rightsretained}

\acmDOI{0000001.0000001}

\received{October 2017}
\received[revised]{September 2018}
\received[revised]{January 2019}
\received[accepted]{February 2019}

\usepackage[american]{babel}
\usepackage[style=american]{csquotes}
\usepackage[inline]{enumitem}
\usepackage{booktabs}

\newcommand{\term}[1]{\textit{#1}}
\newcommand{\principle}[1]{\textit{#1}}

\newcommand{\booktitle}[1]{\textit{#1}}
\newcommand{\abbr}[1]{\MakeTextUppercase{#1}}

\begin{document}


\title[Animation Techniques in HRI User Studies: a Systematic Literature Review]{Animation Techniques in Human-Robot Interaction User Studies: a Systematic Literature Review}

\author{Trenton Schulz}
\author{Jim Torresen}
\author{Jo Herstad}
\affiliation{%
  \institution{University of Oslo}
  \city{Oslo}
  \country{Norway}
}
\email{[trentonw|jimtoer|johe]@ifi.uio.no}

\begin{abstract}
There are many different ways a robot can move in Human-Robot
Interaction. One way is to use techniques from film animation to
instruct the robot to move. This article is
a systematic literature review of human-robot trials, pilots, and
evaluations that have applied techniques from animation to move a
robot. Through 27 articles, we find that animation techniques improves
individual’s interaction with robots, improving individual’s perception of
qualities of a robot, understanding what a robot intends to do, and showing
the robot’s state, or possible emotion. Animation techniques also help
people relate to robots that do not resemble a human or robot. The
studies in the articles show further areas for research, such as
applying animation principles in other types of robots and situations,
combining animation techniques with other modalities, and testing
robots moving with animation techniques over the long term.
\end{abstract}

\ccsdesc[500]{Computer systems organization~Robotic autonomy}
\ccsdesc[300]{Computer systems organization~Robotics}
\ccsdesc[300]{Human-centered computing~HCI design and evaluation methods}
\ccsdesc[300]{Human-centered computing~Interaction paradigms}

\keywords{robot, human-robot interaction, literature review,
animation, motion}

\maketitle

\section{Introduction}

When the Kismet robot was introduced, individuals could interact with it
via conversation or gestures as opposed to typing on a keyboard
\parencite{BreazealHowbuildrobots1999}. Human-robot
interaction (\abbr{hri}) requires the robot to also respond. A robot that
gestures and moves can aid an individual in understanding what the robot is
doing and aid in the interaction.

In movie production, we observed the phenomenon of \term{animation}—layering slightly
different frames of an object to create the illusion of
movement. Animators follow principles such that animations are
believable and tell stories
\parencite{ThomasIllusionLifeDisney1995}. The principles are
successfully used in computer graphics
\parencite{LasseterPrinciplesTraditionalAnimation1987}, and studies
suggested that the principles should be considered for robots
\parencite{vanBreemenBringingrobotslife2004,RibeiroIllusionRoboticLife2012}. However, what is the extent to which
animation techniques are used with robots and how do animation techniques affect \abbr{hri}?

The present study maps the current knowledge by conducting a
systematic literature review of evaluations using animation principles
and techniques in \abbr{hri}. First, we construct a foundation and context
by examining movement, how movement affects an individual’s
interpretation of things, \term{animation} in the \abbr{hri} context, and
animation techniques (Section~\ref{sec:animation}).  Then,
we present the method to perform a systematic review
(Section~\ref{sec:process}). This is followed by the search results
where we provide a review of the articles that we examined
(Section~\ref{sec:results}). We discuss the implications and potential
areas for future research (Section~\ref{sec:discussion}) before
providing a few concluding remarks (Section~\ref{sec:conclusion}).

\section{Background: Movement, animation, and robots}\label{sec:animation}

We first define types of movement. Then, we quickly review principles
of animation as a way of looking at animation techniques for \abbr{hri}
and how they can be applied to robots. We briefly discuss other
techniques for moving robots and conclude the section with an
exploration of the concept of \term{animacy} and its relation to
\abbr{hri} and our study.

\subsection{Movement and Animation}\label{sec:about-movement}

The phenomenon of \term{movement} is straightforward. In physical
terms, movement is a \term{vector} with \term{speed} and
\term{direction}.  In robotics, movement that changes the position of
the robot is called \term{locomotion} or \term{translation}. Robot
movement that does not affect its position is called
\term{configuration}.  Locomotion and configuration can be
combined. So, a robot can move towards a person (locomotion), wave at
a person (configuration), and say “hi”.

Animation in \abbr{hri} uses techniques from animation in films or
computer graphics (or inspiration from them) to specify how a robot
moves. This movement should help a robot communicate with humans. This complements a suggestion by
\textcite{vanBreemenAnimationenginebelievable2004} with using
animation principles to help create “believable behavior”
\parencite[p.~2873]{vanBreemenAnimationenginebelievable2004} in a
robot. \Textcite{RibeiroAnimatingAdelinoRobot2017} built on this
definition and added that “… robot animation consists of all the
processes that give a robot the ability of expressing identity,
emotion and intention during autonomous interaction with human users”
\parencite[p.~388]{RibeiroAnimatingAdelinoRobot2017}.

Let us review some of these animation techniques, starting with the twelve
principles of animation.

\subsection{The Twelve Principles of Animation and Other Animation Techniques}\label{sec:principles}

The idea behind traditional, hand-drawn animations for films corresponds to
physics. That is, switch drawings sufficiently fast such that what is rendered
appears to move. The idea also applies to computer animation or
anything that is filmed. The actual drawing (or rendering) is
considered as art.  \Textcite{ThomasIllusionLifeDisney1995} documented
how animators at Walt Disney Studios practiced their methods of
creating their animations until they obtained a few methods that “…
seemed to produce a predictable result,” \parencite[p.~47]{ThomasIllusionLifeDisney1995}. The artists termed these methods the \term{fundamental principles
of animation}, and the principles were taught to new animators. Although the principles
were not verified scientifically, they have been used in
financially successful animated films and cartoons watched by millions.  There
twelve principles are as follows:

\begin{description}
\item [{Squash and Stretch}] 

Characters and objects should squash and stretch with their action, although they do not completely lose their shape.
\item [{Anticipation}] 

Major action should be telegraphed such as reaching back before throwing an object.
\item [{Staging}] 

An action should be clear to the audience. For example, the audience should understand the action by only viewing it in silhouette. 
\item [{Straight Ahead Action and Pose to Pose}] 

This principle describes how to draw an action. Drawing straight ahead involves starting to draw and simple continuing until the action is completed. Pose to pose implies that specific poses are desired in an action and are choreographed before the actual animation.
\item [{Follow Through and Overlapping Action}] 

Actions are not performed in isolation. An animated character exhibits a plan and moves from one action to the next without stopping between.
\item [{Slow In and Slow Out}] 

The speed of a motion is not the same during the time that it is performed. Action is slower at the beginning and end.
\item [{Arcs}] 

Move limbs in arcs as opposed to of straight up-down and left-right motions.
\item [{Secondary Action}] 

Create complementary actions that emphasize the main action. For example, a character puts on a coat while walking out the door.
\item [{Timing}] 

Changes in number of frames that are between a start and stop determines the speed of the action, thereby increasing the number of frames and decreasing the speed of the action.
\item [{Exaggeration}] 

Exaggerated action ensures that it is easier to understand the feelings of a character.
\item [{Solid Drawing}] 

Drawings should look plausible and three-dimensional and \term{twins}—symmetrical limbs on a character—should be avoided since it makes characters look stiff.
\item [{Appeal}] 

All the characters should be appealing whether one is expected to sympathize with them or despise them.
\end{description}

A few of the principles are related to the craft of pen-and-paper
animation and narrative of films, although they are shown as
applicable to other areas, such as 3-D computer-animated films
\parencite{LasseterPrinciplesTraditionalAnimation1987}.

The twelve principles are not the only methods to animate an object or
produce cartoon-like movement; several other methods reflect aspects of the principles. For example, a common method involves the use of
\term{key frames}, which are frames that define important (key) points in a
movement. Then, the software or other animators interpolate
the frames between the key frames. This is similar to the pose to pose part of the \principle{Straight Ahead Action and Pose to Pose} principle.

A different way of animating movement involves an individual acting out the
movement and transferring it to the animation media. One method is
\term{rotoscoping} where animators trace individual frames of a
filmed action to create a realistic and human-like animation. Another
technique involves the use of \term{motion capture}, where sensors
capture the movement and software translates the movement onto another model.

A field related to animation is \term{puppetry} and
\term{animatronics} where a person controls how a puppet or other
creation moves and reacts to a situation. This is a relevant method to
consider for moving a robot, especially if the robot is
teleoperated. \Textcite{SchererMovieMagicMakes2014} has argued that this is a fertile area to
investigate for robot design.

\term{Kinematics} is a mathematical method to express movement and is
used for robots that are composed of a chain of articulated
nodes. \term{Inverse kinematics} is a method to solve for the
different nodes (joints) to move to obtain a desired position by
working backwards to its starting position. A common use of inverse
kinematics is when a robot arm is picking or placing objects. In the
real world, joints have limited degrees of movement, so not all
solutions are valid. However, applying animation principles to the
formulas (e.g., making movement follow arcs) can turn kinematics into an
animation technique.

\subsection{Other Techniques for Robot Communication Through Movement}\label{sec:non-animation-methods}

Techniques for communicating through movement exist beyond those used in animation
and film. These are not animation techniques, but were developed in
other areas and have been applied to robots.

In the world of dance and acting,
\Citeauthor{LabanModerneducationaldance1948} created the \term{Laban
Effort System} \parencite{LabanModerneducationaldance1948} that
describes human motion in four effort factors: Space, Weight, Time,
and Flow. Each factor has two elements (polarities) to adjust the factor’s
character. For example, Space has elements of direct versus indirect, and Time
has elements of quick versus sustained. The system can be used by dancers and
actors to better understand their own patterns and biases in movements
and impart better quality on their
movement. \Textcite{LaViersStylebasedrobotic2012} used Laban’s work to
make robots dance alongside other dancers using the robots’ own
style. The system was fully formalized for a humanoid robot
\parencite{LaViersStyleBasedRoboticMotion2014}. Knight and her
colleagues implemented a version of the Laban Effort System to
express the internal state of robots with limited degrees of
motion—such as only a head
\parencite{KnightLabanheadmotionsconvey2016} or only a platform that
can turn \parencite{KnightExpressivemotiontheta2014}. They
investigated situations like sharing space in an office environment
\parencite{KnightTakingcandyrobot2015} and putting the Laban System on
top of other tasks the robot was performing
\parencite{KnightLayeringLabanEffort2015}.

Other \abbr{hri} studies have different solutions for robot motion
and communication. Some studies have used colored lights flashing in different
patterns to signify direction
\parencite{SzafirCommunicatingDirectionalityFlying2015} for a flying
drone and what a robot moving in the office is doing
\parencite{BarakaEnhancinghumanunderstanding2016}. Citing an inspiration
from animation, but not necessarily using animation techniques, Dragan
and her colleagues have investigated the difference between what makes a
robot’s motion legible and what makes it predictable
\parencite{DraganLegibilityPredictabilityRobot2013}. This tension
between legible and predictable motion affects collaboration between a
robot and a person \parencite{DraganEffectsRobotMotion2015}. They have
also investigated how a person’s familiarity with a robot affects how
easily the person can predict the robot’s motion
\parencite{DraganFamiliarizationRobotMotion2014}.

\subsection{Animacy}\label{sec:about-animacy}

\term{Animacy} refers to an object moving as if it
is alive (or that it “exhibits life”). The concept was traced back
\parencite{BartneckMeasurementInstrumentsAnthropomorphism2009} to
\citeauthor{Piagetchildconceptionworld1929}’s study of children
learning what is alive or not \parencite{Piagetchildconceptionworld1929}.

The motion that creates animacy is described as \term{animate
motion}: “movement that is self-propelled, but not necessarily created
by other living creatures” \parencite[page~837]{BlakemoreDetectionContingencyAnimacy2003}.  Even simple
shapes can exhibit animacy. In a classic
psychology study by
\textcite{HeiderExperimentalStudyApparent1944}, individuals watched a
film of shapes moving around and then interpreted what happened. A
majority of the individuals described the action in the film as a story and
gave personality traits to the shapes. Subsequently, another study
indicated that individuals perceive animacy in a particle
if it moves on a path and speeds up
\parencite{TremouletPerceptionAnimacyMotion2000}.

Another set of studies examined how
individuals perceived \term{contingency}
\parencite{MichottePerceptionCausality1963}. Individuals watched films of
objects moving and were asked to
interpret them. In a few films, individuals said the movement of one
object (\emph{X}) was contingent on the movement of another object
(\emph{Y}). These aforementioned studies—and studies that built on the
concepts—were reviewed by
\textcite{SchollPerceptualcausalityanimacy2000}.
Another study
used
simple films of objects depicting contingency and animacy to
explore what parts of the brain were activated for each film
\parencite{BlakemoreDetectionContingencyAnimacy2003}.

Several \abbr{hri} studies examined how individuals ascribe feelings and
personalities to the way robots move, whether they look like a dog
\parencite{BatlinerYouStupidTin2004,Bartneckinfluencepeopleculture2006},
a vacuum cleaner
\parencite{ForlizziServiceRobotsDomestic2006,SungMyRoombaRambo2007,SaerbeckPerceptionAffectElicited2010},
or simply an arm \parencite{ZhouExpressiveRobotMotion2017}.  Other
\abbr{hri} animacy studies are based on
\citeauthor{Piagetchildconceptionworld1929} and examine children’s
relationship to robots and other things that are alive
\parencite{MelsonRobotsDogsChildren2005,OkitaExploringyoungchildren2005,BeranUnderstandingHowChildren2011}.
Others have examined how individuals’ interaction with a robot affects
their willingness to end the robot’s existence
\parencite{BartneckKillMockingbirdRobot2007,BartneckDaisydaisygive2007,BartneckDoesDesignRobot2009,Horstmannrobotsocialskills2018}.

Animacy references the original definition of animation (i.e.,
bringing an element to life) and the idea of an animate object—an
object that moves on its own—versus an inanimate object—an object that
does not move. Specifically, animation techniques in Section~\ref{sec:principles} and the other techniques mentioned in Section~\ref{sec:non-animation-methods}
can be used to create animacy. However, this study focuses
on the use of animation techniques and not on animacy generally.

\section{Method: Literature Review Protocol}\label{sec:process}

The systematic review followed a process outlined by
\textcite[p.~1052]{BudgenPerformingSystematicLiterature2006}.
The process consists of five parts:
\begin{enumerate*}[label=(\textit{\alph*})]
\item define a review protocol with research questions and methods employed for assessment,
\item define a search strategy,
\item document the search strategy,
\item specify explicit inclusion and exclusion criteria, and
\item specify the information that will be obtained from each item.

\end{enumerate*}
 We present each part as a subsection here.

\subsection{Research Questions and Methods for Assessment}\label{sec:research-questions}

The goal of the review involved mapping the knowledge that
exists for using animation techniques to move robots and see
where further research can be directed. This resulted in several research
questions:
\begin{enumerate*}[label=(\textit{\alph*})]
\item What animation principles and techniques are used for moving robots?
\item What kind of studies are performed with animated robots and individuals?
\item How do animation techniques affect individual’s interaction with a robot?
\item What data was collected in the aforementioned studies?
\item What robots are used in these studies?
\item What are the environments (lab or real world) in which the studies are conducted?
\item What was the modality for the study (e.g., a live evaluation or a video)?

\end{enumerate*}

Most of the answers are found in the study method, study results, and design
of the robot. So, we can determine candidate articles by searching article metadata.
Then, a reading the method and results section should
determine if the study is relevant for the research questions.

\subsection{Search Strategy Plan}\label{sec:search-strategy-idea}

We followed a similar search strategy employed by
\textcite{RiekWizardOzStudies2012}.  We searched two databases, namley
IEEExplore \parencite{InstituteofElectricalandElectronicsEngineersIEEEXploreDigital2018} and the ACM Digital Library \parencite{AssoctiationofComputingMachinistsACMDigitalLibrary2018}, since they include many articles on
\abbr{hri}, \abbr{hci}, and robotics. Neither databases index the
\abbr{hri} journal the \booktitle{International Journal of Social
Robotics} nor the \abbr{hci} journal \booktitle{Interaction Studies}, but
it is necessary to balance the breadth of the search relative to the complexity
of reproducing the method. The search was performed on 30 June 2018.

The search on IEEExplore only examined metadata, and the search string was as follows:

\texttt{((HRI OR \textquotedbl human-robot
interaction\textquotedbl\xspace) AND (experiment OR \textquotedbl user study\textquotedbl\xspace OR pilot OR evaluation)
AND (animation OR animate OR cartoon))}.

The search of the ACM Digital library searched the
\booktitle{ACM Guide to Computing Literature}
that includes additional items from
other publishers. The search string for the ACM Digital Library was
equivalent to the IEEExplore search string:

\texttt{+(+(HRI \textquotedbl human-robot interaction\textquotedbl\xspace) +(experiment \textquotedbl user
study\textquotedbl\xspace  pilot evaluation) +(animation animate cartoon))}.

We included “cartoon” in the searches since a few studies we were aware
of did not mention animation techniques for movement, but
they mentioned techniques for “cartoon-like movement”.

\subsection{Inclusion and Exclusion Criteria}\label{sec:inclusion-exclusion-criteria}

Beyond the search string, the inclusion
criteria corresponded to peer-reviewed conference and journal articles about robots that used one
or more animation techniques to move and included a study with
individuals. Therefore, a relevant paper included the following:
\begin{enumerate*}[label=(\textit{\alph*})]
\item  at least one robot,
\item at least one animation technique, and
\item at least one person that evaluated or interacted with the robot.

\end{enumerate*}

The goal involved mapping the use of animation techniques in \abbr{hri}
studies, and thus we were generous in what was considered a study and included pilot
studies, informal studies, or critiques of a robot’s movement.

The review excluded posters, workshop announcements, and
non-peer-reviewed books. We also excluded articles that: \begin{enumerate*}[label=(\textit{\alph*})]
\item only described a robot,
\item only described a tool or algorithm for a robot,
\item evaluated robot interaction with animals,
\item only studied animacy (as per Section~\ref{sec:about-animacy}), and
\item only evaluated interaction with virtual agents or virtual robots.

\end{enumerate*}

\subsection{Information obtained from each study}\label{sec:information-obtained}

For each relevant article, we collected information about it for the review. The information was the following:
\begin{enumerate*}[label=(\textit{\alph*})]
\item robot used,
\item embodiment of the robot,
\item animation technique that was used,
\item number of participants,
\item data that was collected,
\item whether the study was performed with a video or in real-life,
\item whether the robot was in a lab or not, and
\item what type of movement was involved (configuration, locomotion, or both).

\end{enumerate*}

\section{Results}\label{sec:results}

The searches returned 68 items from the ACM Digital Library and 46
items from the IEEExplore database. The results from the searches were combined and controlled
for entries that appeared in both the ACM Digital Library and IEEExplore. This resulted in
a total of 106 items (Table~\ref{tab:search-stats}). The searches
produced a sufficient number of articles, although they were not
overwhelming. We began reading the
items to apply the inclusion and exclusion criteria.

\begin{table}[htb]
\caption{Number of articles found in each database.}\label{tab:search-stats}
\center{}
\begin{tabular}{lr}
\toprule
\textbf{Database} & 
\textbf{Results} \\
\midrule

ACM Digital Library & 
68 \\
IEEExplore & 
46 \\
\textbf{In both} & 
(8) \\
\midrule
\textbf{Total} & 
106 \\\bottomrule
\end{tabular}

\end{table}

For articles that matched our inclusion criteria, we wrote down
information as outlined in
Section~\ref{sec:information-obtained}. Articles that were missing
this information or matched our exclusion criteria were excluded, and
the reason for exclusion was documented.

The authors met to discuss the placement of the articles and agreed on a
final list.  We had initial disagreement on six articles \parencite{BreazealEmotionsociablehumanoid2003,YamaokaRelationshipContingencyComplexity2006,YoungRobotexpressionismcartooning2007,TraftonIntegratingvisionaudition2008,RakitaMotionRetargetingMethod2017,DuncanEffectsSpeedCyclicity2017}.
The final consensus was to exclude them as each lacked one of the
inclusion criteria. This resulted in
79 articles that were excluded and
27 that matched the inclusion criteria.

There were three articles we expected to be in the search results, but
they were not in the results due to missing information in the metadata. One
article \parencite{RibeiroIllusionRoboticLife2012} was about applying
animation principles to a robot for showing emotions. The article \emph{does}
include an evaluation, but it is \emph{not} specified in the article
metadata. The second article
\parencite{SzafirCommunicationIntentAssistive2014} used the animation
principles of \principle{Arcs}, \principle{Anticipation}, and
\principle{Slow in and Slow out} for Assistive Free Flying robots, but
there was no mention of animation in the metadata. The third article
\parencite{LuriaDesigningVyorobotic2016} documented the
design process for an animated robot for the smart home but mentioned
neither a user study nor animation in the metadata.  On one hand, it
is unfortunate that the databases missed these articles, and we chose to
keep these specific articles out of the review to keep the method
straightforward to replicate.  On the other hand, several of these
authors \emph{are} included in our list of relevant articles. So,
while a specific article may not be included, their research in this
area is part of the relevant literature.

\subsection{Paper Demographics}\label{sec:demographics}

\begin{table}[htb]
\caption{Breakdown of articles by conference and journal in order of number of articles.}\label{tab:paper-demographics}
\center{}
\begin{tabular}{lp{9cm}r}
\toprule
\textbf{Type} & 
\textbf{Name} & 
\textbf{Articles} \\
\midrule

Conference & 
ACM/IEEE Conference on Human-Robot Interaction (HRI) & 
8 \\
Conference & 
IEEE International Symposium on Robot and Human Interactive Communication (RO-MAN) & 
4 \\
Conference & 
IEEE-RAS International Conference on Humanoid Robots (Humanoids) & 
3 \\
Journal & 
\booktitle{ACM Transactions on Interactive and Intelligent Systems} (TiiS) & 
2 \\
Journal & 
\booktitle{Computers in Human Behavior} & 
2 \\
Journal & 
\booktitle{Autonomous Robots} & 
1 \\
Conference & 
International Conference on Advances in Computer Entertainment Technology (ACE) & 
1 \\
Conference & 
International Conference on Interaction Design and Children (IDC) & 
1 \\
Conference & 
International Conference on Multimodal Interaction (ICMI) & 
1 \\
Conference & 
IEEE Portuguese Meeting on Bioengineering (ENBENG) & 
1 \\
Conference & 
Graphics Interface (GI) & 
1 \\
Journal & 
\booktitle{Journal of Intelligent Robotics Systems} & 
1 \\
Journal & 
\booktitle{Multimedia Tools and Applications} & 
1 \\
\midrule
 & 
\textbf{Total} & 
27 \\\bottomrule
\end{tabular}

\end{table}

The majority of the 27 papers (20) were
conference papers. Over three-quarters of the conference articles
(15) were from \abbr{hri} conferences (\abbr{hri}, RO-MAN, and
Humanoids). The other conferences articles were from conferences that focused on
specialized \abbr{hci} (ACE, IDC, ICMI), graphics (GI), and bioengineering (ENBENG). The remaining seven articles were
from robotics, \abbr{hri}, and \abbr{hci} journals: two journal articles from
\booktitle{ACM Transactions on Interactive Intelligent Systems}
(TiiS); two articles were from \booktitle{Computers in Human Behavior}; and
the last three articles were from
\booktitle{Autonomous Robots}, \booktitle{Journal of Intelligent
Robotic Systems}, and \booktitle{Multimedia Tools and
Applications}. The breakdown of articles from each venue is shown in
Table~\ref{tab:paper-demographics}.

\subsection{Robots and Robot Types Used in the Studies}\label{sec:results-robots}

Although there are articles that examine the use of tools and frameworks that
use techniques from animation for moving a robot
\parencite{vanBreemenAnimationenginebelievable2004,vanBreemenAdvancedAnimationEngine2006,RibeiroNuttyTracksSymbolic2013,BartneckrobotengineMaking2015},
the review examined the animation techniques with robots that are evaluated
with participants, robots that are used in the evaluations.
Table~\ref{tab:robot-animation-technique} sorts the studies by
year and identifies the robot; type of robot (i.e., humanoid, animal, a
head, or other); and animation technique used.

\begin{table}[htbp]
\caption{Studies sorted by year ascending, with robot and animation technique.}\label{tab:robot-animation-technique}
{\footnotesize

\begin{tabular}{rllll}
\toprule
\textbf{Year} & \textbf{Reference} & \textbf{Robot} & \textbf{Type of Robot} & \textbf{Animation Technique}  \\
\midrule
\citeyear{Bennewitzhumanoidmuseumguide2005} & \parencite{Bennewitzhumanoidmuseumguide2005} & Alpha & P & Arcs  \\
\citeyear{YamaokaLifelikebehaviorcommunication2005} & \parencite{YamaokaLifelikebehaviorcommunication2005} & Robovie II & P & Motion Capture  \\
\citeyear{Bartneckperceptionanimacyintelligence2007} & \parencite{Bartneckperceptionanimacyintelligence2007} & iCat, Robovie II & A, P & Secondary Action  \\
\citeyear{DelaunayStudyRetroprojectedRobotic2010} & \parencite{DelaunayStudyRetroprojectedRobotic2010} & RAF & H & Secondary Action  \\
\citeyear{HarrisExploringEmotiveActuation2010} & \parencite{HarrisExploringEmotiveActuation2010} & Stem & O & Unspecified animation techniques  \\
\citeyear{HoffmanInteractiveimprovisationrobotic2011} & \parencite{HoffmanInteractiveimprovisationrobotic2011} & Shinmon & O & Anticipation, Follow Through,  Slow In, Slow Out \\
\citeyear{TakayamaExpressingThoughtImproving2011} & \parencite{TakayamaExpressingThoughtImproving2011} & PR2 & O & Anticipation, Follow Through  \\
\citeyear{WistortTofuDrawMixedrealityChoreography2011} & \parencite{WistortTofuDrawMixedrealityChoreography2011} & Tofu & A & Squash and Stretch  \\
\citeyear{YohananDesignAssessmentHaptic2011} & \parencite{YohananDesignAssessmentHaptic2011} & Haptic Creature & A & Pose to Pose (Key Frame)  \\
\citeyear{BeckEmotionalBodyLanguage2012} & \parencite{BeckEmotionalBodyLanguage2012} & Nao & P & Motion Capture, Pose to Pose (Key Frame) \\
\citeyear{GielniakEnhancingInteractionExaggerated2012} & \parencite{GielniakEnhancingInteractionExaggerated2012} & SIMON & P & Exaggeration  \\
\citeyear{RobertBlendedRealityCharacters2012} & \parencite{RobertBlendedRealityCharacters2012} & Alphabot & O & Pose to Pose (Key Frame)  \\
\citeyear{DawsonItAliveExploring2013} & \parencite{DawsonItAliveExploring2013} & DEVA & O & Squash and Stretch  \\
\citeyear{SharmaCommunicatingAffectFlight2013} & \parencite{SharmaCommunicatingAffectFlight2013} & Parrot AR.Drone & O & Motion Capture  \\
\citeyear{YoungDesignEvaluationTechniques2014} & \parencite{YoungDesignEvaluationTechniques2014} & Roomba, Reactor & O, A & Motion Capture, Puppetry  \\
\citeyear{HoffmanRobotPresenceHuman2015} & \parencite{HoffmanRobotPresenceHuman2015} & Custom Head & H & Unspecified animation techniques  \\
\citeyear{MatthieuArtificialCompanionsPersonal2015} & \parencite{MatthieuArtificialCompanionsPersonal2015} & Nao & P & Pose to Pose (Key Frame)  \\
\citeyear{NitschInvestigatingeffectsrobot2015} & \parencite{NitschInvestigatingeffectsrobot2015} & Nao & P & Secondary Action  \\
\citeyear{ParkGenerationRealisticRobot2015} & \parencite{ParkGenerationRealisticRobot2015} & Custom Head & H & Exaggeration, Secondary Action  \\
\citeyear{YilmazyildizGibberishspeechtool2015} & \parencite{YilmazyildizGibberishspeechtool2015} & Probo & A & Secondary Action, Squash and Stretch  \\
\citeyear{LiSocialRobotsVirtual2016} & \parencite{LiSocialRobotsVirtual2016} & Nao & P & Motion capture  \\
\citeyear{MirnigRobothumorHow2016} & \parencite{MirnigRobothumorHow2016} & Nao, iCat & P, A & Secondary Action, Pose to Pose (Key Frame)  \\
\citeyear{AsselbornKeepmovingExploring2017} & \parencite{AsselbornKeepmovingExploring2017} & Nao & P & Secondary Action  \\
\citeyear{LoureiroISRRobotHeadRobotichead2017} & \parencite{LoureiroISRRobotHeadRobotichead2017} & ISR-RobotHead & H & Secondary Action  \\
\citeyear{RibeiroAnimatingAdelinoRobot2017} & \parencite{RibeiroAnimatingAdelinoRobot2017} & Adelino & A & Inverse Kinematics using animation principles  \\
\citeyear{IzuiImpressionpredictivemodels2017} & \parencite{IzuiImpressionpredictivemodels2017} & Pepper, Nao & P & Pose to Pose (Key Frame) \\
\citeyear{Thimmesch-GillPerceivingemotionsrobot2017} & \parencite{Thimmesch-GillPerceivingemotionsrobot2017} & Nao & P & Puppetry, ``Animation Best Practices''  \\
\bottomrule
\end{tabular}
\caption*{\textbf{Type of Robot}: A: Animal, H: Head, O: Other, P: Humanoid}
}
\end{table}

Twelve studies used a
humanoid robot or a combination of a humanoid robot with an animal
robot. Eight of
the aforementioned studies used Nao
\parencite{BeckEmotionalBodyLanguage2012,MatthieuArtificialCompanionsPersonal2015,NitschInvestigatingeffectsrobot2015,LiSocialRobotsVirtual2016,MirnigRobothumorHow2016,AsselbornKeepmovingExploring2017,IzuiImpressionpredictivemodels2017,Thimmesch-GillPerceivingemotionsrobot2017}
one of the eight also used a Pepper robot
\parencite{IzuiImpressionpredictivemodels2017}. These
commercially-available robots offer software to animate the robot
using animation techniques and using key frames
\parencite{PotChoregraphegraphicaltool2009}. Robovie II is another
commercially available robot that was used for two animation studies
\parencite{YamaokaLifelikebehaviorcommunication2005,Bartneckperceptionanimacyintelligence2007}. Finally,
SIMON and Alpha are custom humanoid robots that were used for one study each
\parencite{GielniakEnhancingInteractionExaggerated2012,Bennewitzhumanoidmuseumguide2005}.

Seven studies
used robots that resembled an animal. Two of the studies used the
iCat
\parencite{Bartneckperceptionanimacyintelligence2007,MirnigRobothumorHow2016},
a cat robot that was designed using animation principles to have an
expressive face
\parencite{vanBreemeniCatAnimatedUserinterface2005}. The other robots
are custom robots. One study
\parencite{WistortTofuDrawMixedrealityChoreography2011} used Tofu, a
fluffy, squash and stretch robot that resembles
a bird. Another study
\parencite{YoungDesignEvaluationTechniques2014} used a plush dog-like
robot to dance. A study
\parencite{YohananDesignAssessmentHaptic2011} used the Haptic
Creature, which resembles a mouse. A study
\parencite{YilmazyildizGibberishspeechtool2015} used Probo, a robot that
resembles a type of mammoth
\parencite{SaldienExpressingEmotionsSocial2010}. The final animal
robot study \parencite{RibeiroAnimatingAdelinoRobot2017} used Adelino,
a custom robot that resembles a snake.

Four studies used
a head to test animation principles. Each robot head was
different. One study
\parencite{DelaunayStudyRetroprojectedRobotic2010} used RAF, a robot that is
a retro-projected face that is projected on a sphere. Another study
\parencite{LoureiroISRRobotHeadRobotichead2017} used the
ISR-RobotHead, a head with LCD screens for the eyes and mouth. Another
study \parencite{HoffmanRobotPresenceHuman2015} used a computer
monitor with animated eyes and neck that moved expressively so that
it was possible to identify where the robot was looking. A robot head with
expressive eyes and a creative use of tubing to make an expressive mouth
was used for the remaining head animation study
\parencite{ParkGenerationRealisticRobot2015}.

Seven
studies used robots that did \emph{not} resemble a animal, head, or
humanoid. These studies represented a variety of robots. Robots
had an appearance of a stick
\parencite{HarrisExploringEmotiveActuation2010}, a large alphabet
block \parencite{RobertBlendedRealityCharacters2012}, or
a smartphone \parencite{DawsonItAliveExploring2013}. Other forms included
domestic robots like the Roomba
\parencite{YoungDesignEvaluationTechniques2014}, a quadcopter drone
\parencite{SharmaCommunicatingAffectFlight2013}, a PR2
\parencite{TakayamaExpressingThoughtImproving2011} or a custom, three-armed,
marimba-playing robot
\parencite{HoffmanInteractiveimprovisationrobotic2011}.

\subsection{Animation Principles and Techniques Used in the Articles}\label{sec:results-principles}

18 studies used one
or more animation principles. This includes counting key frames as a
version of the \principle{Pose-to-Pose}
principle. Some studies explicitly name
the principle. For others, we inferred the
principle from the text, and have noted this below. Table~\ref{tab:animation-breakdown} breaks down the number
of studies for each principle.

\begin{table}[htb]
\caption{Breakdown of animation principle and the number of studies they are used in, ordered by number of articles; some articles use more than one principle.}\label{tab:animation-breakdown}
\center{}
\begin{tabular}{lr}
\toprule
\textbf{Animation Principle} & 
\textbf{Articles} \\
\midrule

Secondary Action & 
8 \\
Straight Ahead Action and Pose to Pose & 
6 \\
Squash and Stretch & 
3 \\
Anticipation & 
2 \\
Exaggeration & 
2 \\
Follow Through and Overlapping Action & 
2 \\
Slow In and Slow Out & 
1 \\
Arcs & 
1 \\
Timing & 
0 \\
Staging & 
0 \\
Solid Drawing & 
0 \\
Appeal & 
0 \\\bottomrule
\end{tabular}

\end{table}

The principle that is most frequently used (eight times) is the principle of
\principle{Secondary Action} where something else is animated in
addition to the main action. The studies that use \principle{Secondary
Action} make the robot react to a situation or show an “emotion” in the acting sense of
showing an emotion as lifeless objects like robots do not have real
emotions
\parencite{Bartneckperceptionanimacyintelligence2007,DelaunayStudyRetroprojectedRobotic2010,NitschInvestigatingeffectsrobot2015,ParkGenerationRealisticRobot2015,YilmazyildizGibberishspeechtool2015,MirnigRobothumorHow2016,AsselbornKeepmovingExploring2017,LoureiroISRRobotHeadRobotichead2017}. In
the aforementioned studies, one \parencite{ParkGenerationRealisticRobot2015} names the principle explicitly and the others imply the principle’s use as they either use a robot that uses this principle (iCat) \parencite{Bartneckperceptionanimacyintelligence2007,MirnigRobothumorHow2016} or document that additional parts are animated during an action (e.g., eyes and eyebrows in addition to the mouth \parencite{DelaunayStudyRetroprojectedRobotic2010,YilmazyildizGibberishspeechtool2015,LoureiroISRRobotHeadRobotichead2017} or moving parts of the body while the robot is idle \parencite{NitschInvestigatingeffectsrobot2015,AsselbornKeepmovingExploring2017}). These secondary actions aid in highlighting what is going on.

The next principle that was used six
times corresponds to \principle{Straight Ahead Action and Pose to Pose}. This principle is
similar to the idea of key frames since—in applying the pose to pose part of the principle—the
animator is trying to create the key poses (i.e., frames) for the
character in a situation. All the studies either explicitly name the method \parencite{RobertBlendedRealityCharacters2012,MatthieuArtificialCompanionsPersonal2015,YohananDesignAssessmentHaptic2011} or use software that uses key poses for driving the animation \parencite{MirnigRobothumorHow2016,BeckEmotionalBodyLanguage2012,IzuiImpressionpredictivemodels2017}.
Studies that employ the principle
examine synchronizing action to another event (e.g., entering or leaving the
virtual world \parencite{RobertBlendedRealityCharacters2012}, dancing
\parencite{MatthieuArtificialCompanionsPersonal2015}, or falling
\parencite{MirnigRobothumorHow2016}), present
the robot’s emotional state
\parencite{YohananDesignAssessmentHaptic2011,BeckEmotionalBodyLanguage2012},
or the impression a participant receives about the robot
\parencite{IzuiImpressionpredictivemodels2017}.

Although most individuals do not consider robots soft and squishy, the
\principle{Squash and Stretch} principle was used in three studies. In two studies
\parencite{WistortTofuDrawMixedrealityChoreography2011,YilmazyildizGibberishspeechtool2015}
the squash and stretch principle was used to make the robot more
appealing to children. Another study, used crawl, breathe, and curl
gestures to create a smartphone that exhibits emotions and appears
alive \parencite{DawsonItAliveExploring2013}. The study does not name
the \principle{Squash and Stretch} principle directly, but the
resulting smartphone and the description of the gestures seem to
evoke it.

The principle of \principle{Exaggeration} was used in two studies such that it was easier for individuals to
understand what the robot was doing. In one study
\parencite{GielniakEnhancingInteractionExaggerated2012}, the SIMON
robot related stories to participants and exaggerated certain gestures
used in the story. The other study
\parencite{ParkGenerationRealisticRobot2015} combined \principle{Exaggeration}
with \principle{Secondary Action} such that it was easier for
participants to understand emotions.

The principles of \principle{Anticipation} and \principle{Follow
Through and Overlapping Action} were used together in two separate
studies to help a non-standard looking robots to express what it was
doing. In one study
\parencite{TakayamaExpressingThoughtImproving2011}, an animator was
employed to design animations following these principles so it was
easier to understand that the robot was delivering a drink, escorting
a person, opening a door, or looking to recharge. One study
\parencite{HoffmanInteractiveimprovisationrobotic2011} used the
aforementioned principles along with the principle of \principle{Slow
in and Slow out} with the marimba-playing robot Shimon to improvise
and signal to jazz musicians playing along with it.

One study had its museum guide robot, Alpha, use sine curves instead
of straight lines to make the robot’s arm movement seem more
human-like \parencite{Bennewitzhumanoidmuseumguide2005}. Although it
is not stated in the article, this is exactly the animation principle
of \principle{Arcs}. The robot’s arms moving in arcs made it easier
for individuals to understand what it was pointing towards.

Rounding out the review of animation principles, a few principles
(\principle{Timing}, \principle{Staging}, \principle{Solid Drawing},
and \principle{Appeal}) are not mentioned in any studies. These
principles have more to do with the craft of creating an animated
film.

With respect to techniques beyond the principles, motion capture was
the most popular other technique and was used in five studies. Two studies used motion capture of
humans as an input to how the robot should react to it. One study
\parencite{YamaokaLifelikebehaviorcommunication2005} used motion
capture to track the robot’s and person’s position. The robot itself
used “nonlinear motion” \parencite[p.~408]{YamaokaLifelikebehaviorcommunication2005}, which could be
interpreted as the \principle{Slow In and Slow Out} principle, but
this is not explicitly specified. Another study
\parencite{YoungDesignEvaluationTechniques2014} motion captured
individual’s movements and used pattern matching and frequency
analysis to generate complimentary trajectories for a Roomba to follow
along and act as the individual’s partner.

The remaining studies used motion capture to capture humans
moving and translate it to robot movement.  In one of these studies
\parencite{BeckEmotionalBodyLanguage2012}, motion captured actors
performing emotions and then used this to animate agents and a
Nao. Another study \parencite{LiSocialRobotsVirtual2016} took videos
of lecturers and converted them to as input for a Nao robot. The final
study \parencite{SharmaCommunicatingAffectFlight2013} motion captured
actors using the Laban Effort System and used this motion to
communicate affect to individuals
using a Parrot AR.Drone.

Two studies used ideas
from puppetry. Puppetry was used as an addition to motion capture as
the second part of a study
\parencite{YoungDesignEvaluationTechniques2014} to teach a robot dog
how to dance by following the movements of a puppet cat. Puppeteers
were consulted along with applying “animation best practices”
\parencite[p.~61]{Thimmesch-GillPerceivingemotionsrobot2017} to creating the
Nao’s body language.

One study \parencite{RibeiroAnimatingAdelinoRobot2017} defined an
inverse kinematics engine such that the Adelino snake robot moved in a
word guessing game. The movements indicated to the human participant
as to how close the participant’s guess was to the correct word.

Finally, two studies
used animation techniques, but the exact method was not
documented. One of the studies
\parencite{HarrisExploringEmotiveActuation2010} cited several
animation techniques and animated movies as inspiration to creating a
concept termed \term{emotive actuation} to move the STEM robot
stick expressively. The other study
\parencite{HoffmanRobotPresenceHuman2015} used animation sketches and
tests to articulate a neck and head such that it appears to be watching
participants.

\subsection{Environments, Participants, Data Collected, Movement Types, and Modality}\label{sec:other-study-details}

After examining the robots and animation techniques used, we
examine other details of the studies.
Table~\ref{tab:robot-environment} shows the studies’ environment (lab
or real world), number of participants, whether the motion was configuration,
locomotion, or both, the data collected, and the modality (video or
live). Given the information, at least 1,180
participants were involved in \abbr{hri} studies that used animation
techniques.

\begin{table}[htbp]
\caption{Studies in same order as Table~\ref{tab:robot-animation-technique} with environment, number of participants, data collected, movement type, and modality.}\label{tab:robot-environment}
\footnotesize

\begin{tabular}{lllp{5.5cm}ll}
\toprule
\textbf{Reference} & \textbf{Setting} & \textbf{\# Participants} & \textbf{Data Collected} & \textbf{Movement} & \textbf{Modality} \\
\midrule
\parencite{Bennewitzhumanoidmuseumguide2005} & Real & Not Listed & Questionnaire: Human-like & C & Live \\
\parencite{YamaokaLifelikebehaviorcommunication2005} & Lab & 23 \& 23 & Questionnaire on cognitive ability, intelligence, Lifelikeness & B & Live \\
\parencite{Bartneckperceptionanimacyintelligence2007} & Lab & 62 & Questionnaire: robot intelligence, animacy & C & Live \\
\parencite{DelaunayStudyRetroprojectedRobotic2010} & Lab & 24 & Where is the robot gazing & C & Live \\
\parencite{HarrisExploringEmotiveActuation2010} & Lab & Not Listed & Design critique, Interpret motion & C & Live \\
\parencite{HoffmanInteractiveimprovisationrobotic2011} & Real, Lab & 6 \& 21 & Hypothesis test, embodiement and appreciation, audience appeal & C & Live \\
\parencite{TakayamaExpressingThoughtImproving2011} & Lab & 273 & Qualitative and rating appeal, intelligence, competence, subordinate & B & Video \\
\parencite{WistortTofuDrawMixedrealityChoreography2011} & Lab & 8 & Observation of children & B & Live \\
\parencite{YohananDesignAssessmentHaptic2011} & Lab & 32 & Questionnaire: pick emotion, SAM, and confidence, plus open questions  & C & Live \\
\parencite{BeckEmotionalBodyLanguage2012} & Lab & 23 & Questionnaire: identify emotion, valence, arousal & C & Live \\
\parencite{GielniakEnhancingInteractionExaggerated2012} & Lab & 54 \& 68 & Test memory of story, test where robot is gazing & C & Live, Video \\
\parencite{RobertBlendedRealityCharacters2012} & Lab & 34 & Qualitative measure for continuity & L & Live \\
\parencite{DawsonItAliveExploring2013} & Lab & 6 \& 10 & Arousal, valence, other things & C & Live \\
\parencite{SharmaCommunicatingAffectFlight2013} & Lab & 18 & Questionnaire: SAM + interview & L & Live \\
\parencite{YoungDesignEvaluationTechniques2014} & Lab & 20, 38, 11 & Observation, Interview & L, C & Live \\
\parencite{HoffmanRobotPresenceHuman2015} & Lab & 60 & Authority, Monitoring, and Guilt & C & Live \\
\parencite{MatthieuArtificialCompanionsPersonal2015} & Real & Not listed & Interest in the set up & C & Live \\
\parencite{NitschInvestigatingeffectsrobot2015} & Lab & 48 & Questionnaire: TA-EG, Competence and enthusiasm, Hypothesis testing & C & Live \\
\parencite{ParkGenerationRealisticRobot2015} & Lab & 25 \& 20 & Compare emotions & C & Video \\
\parencite{YilmazyildizGibberishspeechtool2015} & Lab & 35 & Identify emotion & C & Video \\
\parencite{LiSocialRobotsVirtual2016} & Lab & 40 & Questionnaire: Knowledge recall and attitude, Presentation and enthusiasm & C & Video \\
\parencite{MirnigRobothumorHow2016} & Real & 22 & Questionnaire: Godspeed likability, Big Five Inventory & B & Live \\
\parencite{AsselbornKeepmovingExploring2017} & Lab & 26 & Questionnaire: Godspeed: Perceived Anthropomorphism and Proficiency, Task Performance, and attention & C & Live \\
\parencite{LoureiroISRRobotHeadRobotichead2017} & Lab & 9 & Questionnaire: Identify emotion & C & Video \\
\parencite{RibeiroAnimatingAdelinoRobot2017} & Lab & 42 & Hypothesis testing: Performance, Animation, and Intention & C & Live \\
\parencite{IzuiImpressionpredictivemodels2017} & Lab & 3 & Questionnaire: CH33 (Impression of Robot) & C & Live \\
\parencite{Thimmesch-GillPerceivingemotionsrobot2017} & Lab & 96 & Questionnaire: SAM, robot familiarity & C & Live/VR \\
\bottomrule
\end{tabular}
\caption*{\textbf{Movement Type}: C: Configuration, L: Locomotion, B: Configuration and Locomotion}
\end{table}

\subsection{Video or Live Modalities}\label{sec:study-modalities}

Several \abbr{hri} studies include individuals that interact with a robot in person, while
other studies show a video of the robot performing. Since
the animation techniques are derived from the movie world, it is potentially expected
that most studies use video. However, the opposite was
true since 22 had the study take place
with the participant and the robot in the same setting while only
six used video. Although only six studies used video,
it is possible to recruit many more individuals to look at videos instead
of synchronize a time to meet a robot. They did provide over
one-third of the participants in the studies: 402 participants in
video studies versus 778 participants that interacted with the robot in
person. Most of these 402 participants come from one study
\parencite{TakayamaExpressingThoughtImproving2011} that used Amazon’s
Mechanical Turk to recruit 273 participants. However, with respect to the
median number of participants for video and live (30 and 23
respectively), the number of participants for each study are much
closer.

\subsection{Study Environments}\label{sec:study-environments}

One reason for the literature review was to see how many
studies were done in a lab setting versus studies that were done in a
real world setting. Most of the studies (24)
took place in a lab environment (video modality was counted as a lab
environment). There were four studies
that used an environment outside of the lab (one article
\parencite{HoffmanInteractiveimprovisationrobotic2011} had a study in
a lab and real-world setting for the robot). Two of the studies in the
real world environment
\parencite{Bennewitzhumanoidmuseumguide2005,MatthieuArtificialCompanionsPersonal2015}
did not have a count on the participants or were only a pilot. This
was the case for only one lab study.

\subsection{Studies with Locomotion and Configuration}\label{sec:global-vs-local-movement}

Given the different kinds of movement from
Section~\ref{sec:about-movement}, we wondered what the articles would
say about the movement used in them. Surprisingly, most of the studies
(25) focused on configuration. That is, the robot only moved parts of its body and did not
change its location. Seven studies focused on locomotion. However,
four of the locomotion studies also had the robot do some sort of
configuration (whether it was to shake the person’s hand
\parencite{YamaokaLifelikebehaviorcommunication2005}, squash and
stretch \parencite{WistortTofuDrawMixedrealityChoreography2011},
communicate the robots intention
\parencite{TakayamaExpressingThoughtImproving2011}, or as part of a
humor skit \parencite{MirnigRobothumorHow2016}). Only one study
\parencite{YoungDesignEvaluationTechniques2014} used two different
robots for testing locomotion and configuration.

\subsection{Data Collected and the Affect of Animation Techniques}\label{sec:study-questions}

The studies fall into groups about what researchers were studying: \begin{enumerate*}[label=(\textit{\alph*})]
\item studies where participant should identify the emotion shown by the robot,
\item studies interested in participants’ opinion of a robot’s characteristics,
\item studies asking participants where the robot is looking,
\item studies examining a specific hypothesis for a robot or situation, and
\item pilot studies.

\end{enumerate*}
 The breakdown for the articles is shown in
Table~\ref{tab:study-breakdown}. Let us examine these groups closer.

\begin{table}[htb]
\caption{Breakdown of articles versus what they are studying; some articles appear in multiple categories.}\label{tab:study-breakdown}
\center{}
\begin{tabular}{lr}
\toprule
\textbf{Study Examined} & 
\textbf{Articles} \\
\midrule

Robot emotions & 
9 \\
Robot characteristics & 
9 \\
Specific study hypothesis & 
8 \\
Pilot study & 
3 \\
Robot gaze & 
2 \\\bottomrule
\end{tabular}

\end{table}

Nine
articles looked at interpreting the “emotion” or disposition of the
robot either through the robot’s face or its body language. Of course,
a robot does not have emotions, but it can display expressions that
indicate an emotion. In the studies presented here, there are two main
methods used for assessment. One method has participants rate
the valence (the level of pleasure) and arousal (the level of
enthusiasm) of a robot to create a two-dimensional field of emotion. The other method asks the participant to
identify the robot’s expression as one of the five universal, basic
human emotions as defined by \textcite{EkmanBasicEmotions1999}. These
basic emotions (happiness, sadness, fear, surprise, anger, and
disgust) have corresponding levels of valence and arousal, but may be
easier for individuals to relate to.

Two studies \parencite{BeckEmotionalBodyLanguage2012,DawsonItAliveExploring2013}
asked participant to rate the
valence and arousal using Likert scales
to show that the robots’ movements indicate certain emotions as interpreted by the studies’ participants.
The self-assessment mannequin (SAM)
\parencite{BradleyMeasuringemotionselfassessment1994} offers a
alternative method using only pictures for identifying arousal and
valence, and creates similar results.  The SAM was used in three
articles in the review
\parencite{YohananDesignAssessmentHaptic2011,SharmaCommunicatingAffectFlight2013,Thimmesch-GillPerceivingemotionsrobot2017}. One
study used the SAM with the Haptic Creature
\parencite{YohananDesignAssessmentHaptic2011} and found that the
robot’s motion communicated four of the nine conditions correctly to
participants, and participants had correctly identified arousal
correctly, but less well the valence. The second study
\parencite{SharmaCommunicatingAffectFlight2013} had statistically
significant results for valence and arousal in the Laban Effort System factors of
Space, Weight, and Time, but only for arousal for the factor of Flow. The third study
using SAM \parencite{Thimmesch-GillPerceivingemotionsrobot2017} showed
that the valence and arousal of the robot’s movements were reduced
when the person was under a stressful condition.

The method for using \citeauthor{EkmanBasicEmotions1999}’s basic emotions is to ask participants
to look at the robot and pick the corresponding emotion. The final
results are then compared against the chance of someone randomly
picking emotions. Some articles that were excluded had participants match the facial expression using static
pictures of robots (e.g.,
\parencite{BreazealEmotionsociablehumanoid2003,SosnowskiEDDIEEmotionDisplayDynamic2006,DanevDevelopmentanimatedfacial2017}),
but four articles in the review
\parencite{YohananDesignAssessmentHaptic2011,ParkGenerationRealisticRobot2015,YilmazyildizGibberishspeechtool2015,LoureiroISRRobotHeadRobotichead2017}
ran the evaluation with robots that were animated and used secondary
action. Regardless of if the robot was animated or not, the
selections of the participants matched the shown emotion well above
chance, especially for happiness or sadness. But participants showed
confusion between some other emotions (e.g., disgust was often misidentified
as anger).

The nine articles evaluating
characteristics of the robot were concerned with the participants’
opinion about the robots motion or other qualities.  The earliest
study \parencite{Bennewitzhumanoidmuseumguide2005} asked individuals
visiting their stand how human-like the robot’s arcing arm motions
were, with the arcs generally making the motion appear to be similar
to humans. One of the questions in another study
\parencite{GielniakEnhancingInteractionExaggerated2012} was for individuals
to classify how different amounts of exaggeration in the robot’s
motion yielded more cartoon-like or human-like movement. A different
study \parencite{YamaokaLifelikebehaviorcommunication2005} looked at
lifelikeness, but also asked about the robot’s cognitive ability and
intelligence. The robot scored higher when its motions were reactive
of the person interacting with it, than if the motions were simply
static. This measurement was further developed in a later study
\parencite{Bartneckperceptionanimacyintelligence2007} to include
animacy, where participants worked with either a Robovie II or an iCat
to play a game. Though participants found Robovie II to be more
intelligent than the iCat despite them both giving similar advice,
participants spent more time looking at the iCat’s animated face than
they did the Robovie. A different study
\parencite{TakayamaExpressingThoughtImproving2011} had participants
rate the robot’s appeal, intelligence, competence, and how subordinate
it was on a Likert scale along with describing what was happening in
the scene. Here, the robot that was animated to show forethought
before it did a task increased its appeal. Similarly, a robot that
reacted to succeeding or failing a task made participants feel that
the robot had intelligence and competence. As part of another study
\parencite{LiSocialRobotsVirtual2016}, participants were asked to rate
a lecturer’s likability and attitude for delivering a video
presentation with most participants preferring the human form or an
animation using the same voice over a robot or an animation of a
robot.

The Godspeed Questionnaire
\parencite{BartneckMeasurementInstrumentsAnthropomorphism2009} was
created as a standard way to evaluate participants’ perceptions of different aspects of a robot
interaction. The questionnaire consists of scales for
Anthropomorphism, Animacy, Likeability, Perceived Intelligence, and
Perceived Safety. Each scale is independent, so \abbr{hri} researchers can
choose the relevant scales that work for them. The questionnaire shows
up in two articles in this review
\parencite{MirnigRobothumorHow2016,AsselbornKeepmovingExploring2017},
One article \parencite{MirnigRobothumorHow2016} looked at Likability
between two robots and showed how a robot could improve its likability
by laughing at itself after it fell over. The other study
\parencite{AsselbornKeepmovingExploring2017} used the Anthropomorphism
and Proficiency scales to compare two robots, one moving only for
static situations and one moving when it was idle, the idle action robot
attracted more attention and scored higher on the anthropomorphism
scale. A separate method for evaluating safety and performance
qualities of robots, the CH33, was developed in Japan
\parencite{KamideNewMeasurementPsychological2012} and was used in one
study in this review \parencite{IzuiImpressionpredictivemodels2017} to
examine how well a model of motion perception matched to the
perception of individuals watching different types of robot motion.

Eight
articles had a specific hypothesis that was being tested.  One
article \parencite{HoffmanInteractiveimprovisationrobotic2011}
investigated the musicians’ appreciation for seeing the robot’s motions
when they improvised with it and how much having the robot and
musician on stage appealed to the audience watching. A different study
\parencite{HoffmanRobotPresenceHuman2015} examined the feelings of a
person doing a task with an animated robot watching. Though
participants could cheat for a better result in their task, they
tended to be more honest with the robot watching with possible
negative attitudes towards the robot.  Another study
\parencite{GielniakEnhancingInteractionExaggerated2012} found that the
exaggerated motions of the robot storyteller made those parts of the
story more memorable.  Another study
\parencite{NitschInvestigatingeffectsrobot2015} used animation
techniques to simulate competence and enthusiasm in a robot playing
the ultimatum game with a
participant. \Textcite{RibeiroAnimatingAdelinoRobot2017} had
participants rate the performance of the robot, its animation, and its
intention. A different study \parencite{LiSocialRobotsVirtual2016}
looked at how much each student remembered from each lecture from a
human, an animated human using the lecturer’s voice, a robot, and an
animation of the robot. The human lecturer followed by the animation
of the robot resulted in the best scores for the participants’
knowledge.

Two
studies used only qualitative methods.  One study
\parencite{RobertBlendedRealityCharacters2012} asked qualitative
question about what children thought of the Alphabot and how the
children understood the robot entering and leaving the virtual
world. The other study \parencite{YoungDesignEvaluationTechniques2014}
used observation and interviews to find out which methods worked best
for teaching robots new ways to move.

There were two studies that used
animation techniques and investigated where participants thought the
robot was looking.  One study
\parencite{DelaunayStudyRetroprojectedRobotic2010} compared gaze
direction with a spherical robot head versus a flat screen
monitor. The spherical shape of the head and its use of secondary
action in its eyes made it easier to see what was being looked at than
the flat screen monitor.  The other study
\parencite{GielniakEnhancingInteractionExaggerated2012} showed that
the exaggerated motion of the robot made it easier for participants to
predict the direction of the eye gaze than if the robot’s motion
wasn’t exaggerated.

Finally, there were three studies that
tested an animation technique with some participants to see if a
concept could be further developed.  Two studies
\parencite{MatthieuArtificialCompanionsPersonal2015,WistortTofuDrawMixedrealityChoreography2011}
involved testing if a specific set up would work with children, with
general success.  The other study in this group
\parencite{HarrisExploringEmotiveActuation2010} was a design critique
of a stick robot and how it moved.

\section{Discussion}\label{sec:discussion}

This systematic review has looked at \abbr{hri} studies done with robots
that move using techniques from animation. What do these articles say
about this area of research and what are future directions for research?

\subsection{The Articles as a Whole}\label{sec:articles-as-a-group}

Table~\ref{tab:robot-animation-technique} shows that there have been
some \abbr{hri} studies using animation techniques back in the mid-2000s
and at least one article about animation techniques in an \abbr{hri} study
every year since 2010. So, researchers are interested in researching animation
techniques and robots and see how it affects individual’s interaction with
the robot.

Animation techniques help a robot
communicating with a person, either directly or indirectly.  Motion
from animated techniques can make it easier to express some
emotions. Animation techniques also help making a robot appear more
appealing to the individuals who are either watching the robot or
interacting with it. It can make the robot easier to relate to,
approachable, or to have more intelligence.

The studies also show that animation techniques help beyond
communicating an emotion. Motion from animation techniques can draw
individual’s attention to the robot.  It can aid in understanding where a
robot is looking, what it is planning on doing, or going to do next.
This makes it easier to cooperate for human and robots to work
together on a shared task.

The studies also indicate that animation techniques are useful for
robots that do not have a standard animal or humanoid form.
\Textcite{HoffmanDesigningRobotsMovement2014} suggest that robot
forms that are different from animals and humanoids may need to move
in ways that are familiar to individuals to help individuals understand the
robot. Animation techniques provide a method of movement that is
familiar to individuals and easy to relate to based on the nearly a century
of animation techniques in other media.

Looking at Table~\ref{tab:robot-environment}, we can see there are
good measurement tools available for looking at aspects of using
animation techniques with robots and comparing with other
studies. This can help connect new research in animation techniques to
the already existing research. If using an animation technique is to
make the robot appear more likeable, safe, alive, or intelligent than
the Godspeed questionnaire is a readily available measure that has
been used by studies using animation and other studies
\parencite{WeissMetaanalysisusage2015}. It can be a useful tool to
compare new research with past results. If the goal of a study with
animation techniques is to convey emotions, either using the basic
emotions of \textcite{EkmanBasicEmotions1999}, SAM, or rating valence
and arousal provide a way of comparing results with past studies using
other movement techniques. Of course, other qualitative and
quantitative methods can be applied to look at new areas.

In general, the studies seem to indicate that using animation
techniques is overall a positive experience for the individuals interacting
with the robot. Returning to
\citeauthor{RibeiroAnimatingAdelinoRobot2017}’s definition
\parencite{RibeiroAnimatingAdelinoRobot2017} from
Section~\ref{sec:about-movement}, animation techniques can certainly
help make robots’ behavior believable and allow robots to express
identity, emotion, and intention. This suggests that spending time
thinking about how a robot’s motion will be perceived by others should aid in creating better robots to interact with, especially if
robots may be part of what we see in our future everyday lives. Designers and engineers can enlist the support of animators,
puppeteers, and others for determining how a robot should move (e.g.,
\parencite{LuriaDesigningVyorobotic2016,Hoffmanhybridcontrolsystem2008,SchererMovieMagicMakes2014}).

\subsection{Future Research Directions}\label{sec:future-work}

This literature review also points to different areas where further
research in using animation techniques with \abbr{hri} studies. These are
some possibilities.

The twelve principles of animation are an area that can be further
explored. Table~\ref{tab:animation-breakdown} shows that four of the
twelve had no study related to them. Some of these principles, like
\principle{Staging} and \principle{Timing}, may seem to apply only for
framing and directing a movie, but even bits of these principles may
be still be applicable to robots.  For example, the principle of
\principle{Staging} states that action should be understandable only
by watching the silhouette, and this could aid individuals checking the
robots action from a distance. Even the principles that are about
aesthetics (\principle{Solid Drawing} and \principle{Appeal}) are
useful for creating motion for robots (avoiding symmetrical motion or
stopping of limbs) or designing a robot (making the robot appealing to
individuals who will be interacting with it).

\principle{Secondary Action} is used in several articles to add a
small animation to help convey another action.  But it was mostly used
for humanoid or head robots, and the one animal robot, Probo, has a
more human-like face. It would be interesting if this could also be
applied to the non-human, non-animal robots. For example, a part on
the non-humanoid, non-animal robot on could be animated to have an
analog of a blink.

Other principles can also be investigated on other types of
robots. For example, the principle of \principle{Slow in and Slow out}
is only used in one study here, but it could likely be employed in
many situations of different types of robot motion. The
principle of \principle{Arcs} could also be used for other types of
robot motion. The \principle{Squash and Stretch} principle can
pose an interesting challenge to individual’s assumptions of a robot made of hard materials.

Another principle that could be looked at is the principle of
\principle{Follow Through and Overlapping Action}. One obvious place
is the transition from configuration to locomotion or when
locomotion and configuration are combined. This would also be an
opportunity to examine more of the animation principles using locomotion.

Since animation techniques have been adapted in computer animation
\parencite{LasseterPrinciplesTraditionalAnimation1987}, they have also
shown up in graphical user interfaces on computers
\parencite{ChangAnimationCartoonsUser1993,HudsonAnimationSupportUser1993}. So,
some of these techniques have already been formalized. This is another
area where tools used for creating computer animation and games can be adjusted
to work with robots \parencite{BartneckrobotengineMaking2015}.

Using formalization from animation techniques to computer algorithms
from above, animation techniques may also be a way of achieving motion
that is defined in other ways. For example,
\textcite{LaViersStyleBasedRoboticMotion2014} and
\textcite{KnightLayeringLabanEffort2015} worked on formalizing the
Laban Effort System for different robots. One study in the review
\parencite{SharmaCommunicatingAffectFlight2013} provides an example of
using the animation techniques of motion capture to demonstrate how to
move a drone as expressed via the Laban Effort System.

Animation techniques could also aid in the combating the \term{uncanny
valley} (re-translated to English as
\textcite{MoriUncannyValleyField2012}). The uncanny valley is the idea
that there exists a curve representing an individual’s affinity towards a robot
versus how human-like the robot looks. As the robot looks more
human-like, the individual’s affinity grows until it peaks and suddenly
the looks are \emph{not good enough} (i.e., uncanny) and the individual’s affinity for the robot wanes.
Continuing through the valley, at some point the robot’s looks near that of a human
and the individual’s affinity for it rises again.

Although the uncanny valley is focused on the robot’s looks,
\citeauthor{MoriUncannyValleyField2012} posited that more
machine-like movement than organic movement makes the slopes in the
valley even steeper. That is, if something \emph{looks} more like a
human, but does not \emph{move} like a human, then it is difficult for
us to have affinity for it.
\Citeauthor{TakayukiKandaHumanRobotInteractionSocial2012} claimed
that a robot that resembles a human, but does not move like one is
“unnatural” \parencite[p. 101]{TakayukiKandaHumanRobotInteractionSocial2012}.  Since
animation techniques affect how things move, they could also help in
addressing this. Some articles in the review
\parencite{MirnigRobothumorHow2016,LoureiroISRRobotHeadRobotichead2017} mention the uncanny valley
explicitly as a motivation for their research.

Note that animation techniques do not solve all problems. Animation
that is created to be shown on a screen is free of limitations of the
physical world. Servos and other methods for movement have limitations
in strength, friction, flexibility in movement, and other
issues. These limitations need to be considered if an animation
technique will move from the screen to a robot. But this is another
area that could be explored: the quality of the animation created by
the animation techniques and how this affects interaction. That is,
what separates good animation from bad animation in robots? This may
be useful if other considerations such as limited movement or energy
conservation must be balanced against interaction with the robot.

Future research could look at the combination of animation
techniques with the other modalities like sound or smell. This may
result in a stronger or weaker effect than just the animation technique
alone. Combining modalities also makes the robot more universally
designed and accessible to more individuals. A robot moving its limbs to
communicate its intention is useless if the individuals it is interacting
with cannot see it.

Most of the studies in this review took place in a lab setting with
one-on-one interaction. Even though a lab provides an environment to
ensure a robot work well, others have advocated that it is important
to try to get \abbr{hri} studies out into real-world settings and test
interaction over a longer term
\parencite{JungRobotsWildTime2018,DautenhahnBriefThoughtsFuture2018}.
Testing robots in the real-world will help determine how well motion
using animation techniques works when competing or cooperating with
other elements in the environment, and if the animation is effective
or annoying over long term exposure. This may also mean not using
video recordings of the robot and instead focus on individuals working with
the robot live.

Having studies that take place outside of the lab also allow the
introduction of non-lab contexts. One psychology study shows that
context can affect how individuals perceive human faces
\parencite{RighartContextInfluencesEarly2006}. Further research is
needed to see if context has an effect on how individuals perceive robots’
faces and actions.

Although there were some methods that showed up multiple times (e.g.,
the Godspeed Questionnaire, SAM, and choosing from Ekman’s basic
emotion), future researchers should not feel that these are the only
methods that can work for evaluating animation techniques in
\abbr{hri}. Other methods also exist for evaluating the emotion a robot is
displaying, such as the circumplex model of affect
\parencite{Posnercircumplexmodelaffect2005}. Quantitative methods testing a
hypothesis were used in several studies and may fit for certain
studies.  Furthermore, in some situations, such as working with
children or looking for a deeper understanding of a phenomenon,
qualitative observations and interviews are necessary.

Finally, this review has focused on the use of animation
techniques. As mentioned in Section~\ref{sec:about-animacy}, animacy
is a closely related concept and animation techniques can certainly
lead to the perception of animacy in a robot, though it is not the
only way this can be done. There was some effort involved in separating
articles out about animation technique and the concept of
animacy. With this review of animation techniques in \abbr{hri} studies
completed, it makes the task of looking at animacy in \abbr{hri} studies
more straight forward.

\section{Conclusion}\label{sec:conclusion}

We have run a systematic review animation techniques from movies and
computer animation in user studies and evaluations in \abbr{hri}. This
resulted in 27 out of a total of
106 articles that were returned from the ACM
Digital Library and IEEExplore. There have been several animation
techniques that have been adapted to work with \abbr{hri}; this includes
researchers using the twelve principles of animation
(Section~\ref{sec:principles}) and other techniques like motion
capture. The studies in the articles show that motion created through
animation techniques affect an individual’s impression of the robot, help the
robot express intention, or help individuals understand an expression a
robot is showing. Having a better understanding of a robot can make it
easier to interact with a robot, and it can also make it easier for
the robot to interact with individuals.

The literature has shown that animation techniques can help in \abbr{hri}
and is an area that can be further researched. Given that animation
techniques help in the motion of a robot, they are applicable in
different types of \abbr{hri} studies. If a researcher is interested in
making a robot move distinctively to help interaction,
animation techniques are good places to investigate.

There is much to discover about animation techniques, robots, and
\abbr{hri}. Future researchers have a fertile frontier to explore
in helping humans and robots interact better together.

\begin{acks}
The authors thank Tone Bratteteig, Guri Birgitte Verne, Jorunn B{\o}rsting,
Bjarte {\O}stvold and all the others who hove read through
earlier revisions of this article and provided valuable comments. They also thank
the reviewers who provided excellent advice and suggestions for clarity to the manuscript.

This work is part of the \abbr{mecs} project supported by the
\grantsponsor{nfr1000}{Norwegian Research Council}{https://www.nfr.no}
under Grant No:~\grantnum{nfr1000}{247697}
of the~\grantnum[https://www.forskningsradet.no/prognett-iktpluss/Home_page/1254002053513]{nfr1000}{IKTPluss
Program}.
\end{acks}

\printbibliography

\end{document}